# Agentic AI platforms for autonomous training and rule induction of human-human and virus-human protein-protein interactions

Hung N. Do[1,*], Jessica Z. Kubicek-Sutherland[2], Oscar A. Negrete[3], and S. Gnanakaran[1,*]

[1]Theoretical Biology and Biophysics Group, Theoretical Division; [2]Physical Chemistry and Applied Spectroscopy Group, Chemistry Division, Los Alamos National Laboratory, Los Alamos, New Mexico, USA 87545

[3]Systems Biology Department, Sandia National Laboratories, Livermore, California, USA 94550

*Correspondence: gnana@lanl.gov and hungd238@lanl.gov

## Abstract

Artificial intelligence (AI) has made significant advancements in recent years, helping accelerate the development of solutions to challenging problems. Protein-protein interactions (PPI) govern crucial biological processes and drive systems biology, medical countermeasures, and mechanistic host-pathogen research. However, PPI prediction remains challenging not only because of incomplete positive datasets but also because of the lack of negative data labels. Here, we instruct an AI agent to construct two separate agentic AI platforms: one for autonomous training of predictive ML models for human-human and virus-human PPI, and the other for inducing explicit general rules governing human-human and virus-human PPI. The first agentic AI platform for autonomous training of predictive ML models for PPI is designed to consist of five AI agents that handle autonomous data collection, data verification, feature embedding, model design, and training and validation on three-way protein-disjoint cross-fold datasets. For human-human and human-virus PPIs, the final three-way protein-disjoint ensemble achieves an accuracy of 87.3% and 86.5%, respectively. For cross-checking and interpretability purposes, the second agentic AI platform is designed to replace ML predictions with human-readable rules derived from protein embeddings, physicochemical autocovariance descriptors, compartment annotations, pathway-domain overlap, and graph contexts. For human-human PPI, it is defined by a two-rule induction, whereas human-virus is induced by a more complex set of weighted rules. The rules induced by the second agentic platform align with the SHAP-identified features from the predictive ML models built by the first agentic platform. Taken together, our work demonstrates the agentic AI's ability to orchestrate from data planning to execution, and from rule induction to explanation in ML, opening the door to various applications.

## Introduction

Artificial intelligence (AI) has advanced significantly in recent years. The most recent AI technology is AI agents, which have been applied to solve various challenging problems in biology[1-5]. In particular, Swanson et al. built a virtual lab of AI agents consisting of a large language model (LLM) principal investigator guiding a team of LLM scientist agents to design novel nanobodies for SARS-CoV-2[1]. Alber et al. came up with CellVoyager, a computational biologist AI agent, to autonomously analyze scRNA-seq and demonstrated that AI biologist could uncover new insights in COVID-19 cell-cell communications and aging[2]. Furthermore, Huang et al. developed Biomni, a general-purpose virtual biomedical AI biologist capable of carrying out tasks across genetics, genomics, microbiology, pharmacology, and clinical medicine[3]. More recently, Zhang et al. even built a virtual biotech, led by a Chief Scientific Officer agent and composed of a team of domain-specialized scientist agents, for therapeutic discovery and development[4].

Protein-protein interactions (PPI) govern crucial biological processes that drive daily functions of organisms. The first computational work for predicting PPI used a mixed Bayesian model as the predictor[6,7]. Other studies relied on interolog transfer, domain-domain, and motif-domain interference to make predictions[8,9]. Shen et al. applied the support vector machine (SVM) algorithm with sequence-derived features to predict human PPI[10]. Later, Zhang et al. introduced PrePPI, the first genome-scale structure-based predictor for human PPIs[11], which was later expanded by Garzon et al. into a computational interactome for ~85% of the human proteome[12]. DeNovo was developed to predict PPIs for novel viruses with little to no interaction data[13]. Furthermore, D-SCRIPT was developed as a sequence-based, structure-aware deep learning (DL) to predict PPIs[14]. Despite the large number of studies[15-20], machine learning (ML) predictions of PPIs remain difficult due to the scarcity of positive data and the lack of experimentally confirmed negative data. Furthermore, several ML models exhibited highly inflated performance due to data leakage between the training and validation datasets[20]. These challenges are magnified when viral proteins are involved in PPI.

In this work, we explored the possibility of using artificial intelligence (AI) agents to develop autonomous agentic AI platforms for training and developing ML architectures to predict human-human and virus-human PPIs, as well as for the induction of biological rules governing these PPIs (**Figures 1-2**). A coding agent received instructions from us to develop the two agentic AI platforms. The first agentic AI platform handled all steps involved in training predictive ML models autonomously, including data collection, data verification, feature embedding, model design, training, and validation, as well as the explanation of predictions using the SHAP (Shapley additive explanation[21] approach (**Figure 1**). The second agentic AI platform was developed based on the first agentic platform by replacing the training and validation tasks with rule induction to produce explicit biological rule sets for PPI, serving as a cross-check for the predictive models and as an interpretation of ML results (**Figure 2**). Our work illustrated the agentic orchestration of ML training and the coupling of predictive models with scientific interpretability within reproducible computational workflows.

**Methods**

We constructed two agentic AI platforms for autonomous training of predictive ML models (**Figure 1**) and rule induction (**Figure 2**) of human-human and virus-human PPI. We provided high-level instructions to a coding agent, which was ChatGPT 5.2 and 5.4 Thinking, to design the two agentic AI platforms. Each agentic AI platform was implemented in a separate *Python* script, with the agentic AI platform for autonomous training of predictive ML models (**Figure 1**) developed first, and the agentic AI platform for rule induction of PPI (**Figure 2**) developed from it. In total, the agentic AI platform for autonomous training of predictive ML models for PPI comprised five AI agents with specified functions: data collector, data verifier, feature embedder, model designer, and executor (**Figure 1**). On the other hand, the agentic AI platform for rule induction of PPI consisted of four AI agents, retaining three of the five AI agents from the platform for autonomous training, including a data collector, data verifier, feature embedder, and replacing the model designer and executor with the rule induction agent (**Figure 2**). All AI agents in the platforms were based on OpenAI API gpt-5.4 models. The developments and executions of the agentic AI workflows were carried out on a 2023 MacBook Pro with Apple M2 Max chip and 32GB memory, as well as an AMD Ryzen Threadripper PRO 5975WX 32-core workstation running on a Linux x86_64 operating system equipped with 64 CPUs and four NVIDIA RTX 6000 Ada Generation graphics cards with 49 GB GPU memory each.

*Agentic AI platform for autonomous training of predictive ML models for human-human and virus-human PPI*

Five AI agents with specified functions comprised the agentic AI platform for autonomous training of predictive ML models for PPI, resembling the process of human training of ML models (**Figure 1**). The first AI agent, the data collector, harvested the positive and negative datasets based on reliable publications, databases, and developer-provided, manually curated datasets (**Figure 1**). In particular, the data collector agent scouted the literature through Europe PMC (https://europepmc.org/) queries followed by best-effort retrieval of abstracts, full texts, and supplementary files. Furthermore, for human-human PPI, the data collection agent targeted the IntAct[22] and BioGRID[23,24] databases and incorporated developer-provided datasets collected from Richoux et al.[25,26] for the collection of positive data (**Supplementary Data 1**). Meanwhile, the negative data were retrieved and synthesized from developer-provided datasets collected from Richoux et al.[25,26], automated Negatome[27,28] retrieval, compartment-based exclusion, and CORUM[29]-informed co-complex filtering (**Supplementary Data 1**). For the virus-human PPI, the data collector agent primarily collected data from HVIDB[30], VirusMentha[31], VirHostNet[32,33], and HPIDb[34,35] for the positive datasets (**Supplementary Data 2**), while the negative data were primarily directly synthesized by consulting with the limited data available in publications, Negatome[27,28], and CORUM[29] , and excluding known interactions (**Supplementary Data 2**). The sequences in the positive and negative datasets for human-human and virus-human PPI were retrieved from UniProt[36] by the AI agent (**Figure 1**) and split into training, validation, and test sets at a 70:10:20 protein ratio, rather than interaction pairs, to minimize data leakage. In particular, any pairs containing held-out test proteins were placed in the test sets, any remaining pairs containing held-out validation proteins were included in the validation sets, and only pairs containing neither testing nor validation proteins were placed in the training sets.

The curated datasets by the data collector agent were then provided to the second AI agent, namely the data verifier agent, to check for possible data hallucinations, over-representations of proteins in both positive and negative datasets as well as the training, validation, and test sets (class balance and negative inflation), and possible overlapping of proteins within the training, validation, and test sets (data leakage risks) (**Figure 1**). If any violations were detected by the data verifier agent, feedback would then be provided to the data collector to re-curate and rebuild datasets (**Figure 1**).

If the verification of the curated data was successful, the verified data would then be supplied to the feature embedder agent, the third AI agent in the platform (**Figure 1**). Afterwards, the feature embedder agent turned the protein amino acid sequences into multimodal representation spaces made of protein language-model embeddings[37,38] and physicochemical autocovariance descriptors[39,40] (**Figure 1**). Here, a configurable protein language model (PLM), with ProtBert[37] as the default encoder and evolutionary-scale modeling (ESM)-family models[38] with tokenization as alternative options, generated pooled embeddings from protein amino acid sequences (**Figure 1**). The embeddings created by the PLM were concatenated with the physicochemical autocovariance descriptors of five-dimensional (5D) Sandberg z-scale (Z-ACC) [40-42] and one-dimensional (1D) Eisenberg consensus hydrophobicity scale (E-ACC)[39,41,42] computed over the protein sequences at configurable lag values (**Figure 1**). Here, the 5D Sandberg z-scales refer to the principal-component descriptors derived from a large set of amino-acid physicochemical measurements, with each amino acid encoded by a five-number vector[40]. In particular, the first z-dimension (z1) tracks hydrophobicity or lipophilicity, z2 determines size along with steric bulk and polarizability, z3 monitors polarity and charge-related behavior, while z4 and z5 extend the representation into higher-order electronic and reactivity-related directions involving electronegativity, heat-of-formation behavior, electrophilicity, or hardness[40]. Meanwhile, the 1D Eisenberg consensus hydrophobicity scale assigns a single scalar to each amino acid and is a compact descriptor of residue hydrophobicity and hydrophilicity[39]. Furthermore, the lagged auto-covariance transformation (ACC) of residue-wise descriptors was employed to describe how residue properties co-varied across sequence separation, which was useful for alignment-free descriptors for downstream ML[41,42]. In particular, ACC mean-centered either the 1D Eisenberg scalar or 5D Sandberg vectors for each amino acid across one amino acid sequence and then computed the average element-wise product between residues $i$ and $i+lag$[41,42]. The default lag value was five, and the employment of ACC converted variable-length sequences into fixed-length summaries that preserve short- to medium-range sequence-order dependence[41,42].

As described, the sequence descriptors were concatenated with PLM embeddings in the form of *concat(proteinA, proteinB, |proteinA – proteinB|, proteinA x proteinB)*. The construction preserved information from both proteins while adding an explicit contrast channel via the absolute difference of their feature vectors, as well as a concordance channel via the element-wise product. In fact, the construction exposed both protein-specific and interaction information to downstream ML models.

The embedded features were then provided to the fourth AI agent in the agentic AI platform, the model designer (**Figure 1**). The model designer agent explored different ML classifier options, including multilayer perceptrons (MLP), two-tower architectures, pair transformers, token transformers, and cross transformers (**Figure 1**). Furthermore, the AI agent selected between binary cross-entropy or focal for loss functions, while deciding between rectified linear unit (ReLU)[43], Gaussian error linear unit (GELU)[44], hyperbolic tangent function (tanh)[45], and sigmoid linear unit (SiLU)[46] for activation functions of MLP and two-tower models, and GELU[44] or ReLU[43] for the encoded layers in the transformers. The described ML architectures were included in the search space for the model designer agent, as they are typical and encompass a wide range of ML scenarios. In particular, the MLP is one of the simplest ML models, operating directly on flattened pair vectors. Meanwhile, the pair transformer treats *concat(proteinA, proteinB, |proteinA – proteinB|, proteinA x proteinB)* as four-token sequences and learns pair vectors through self-attention. The token transformer treats *concat(proteinA, proteinB, |proteinA – proteinB|, proteinA x proteinB)* as either a two-token or a four-token sequence, with alternative pooling strategies. The cross-transformer is a light fusion encoder that lets the two proteins exchange information without rigid four-token joint encoders. Lastly, the two-tower model encodes the two proteins separately into latent vectors and then scores them by a temperature-scaled similarity function, with an optional small commutative correction head (**Supplementary Figure 1**).

The process of model selection was rather empirical than rule based. First, the model designer agent first proposed candidate model specifications. For MLP, the default architecture was made of three layers with hidden sizes of 512, 256, and 128. For transformers, the default transformer encoders consisted of two layers with the dimensionality of input and output representations being 512 and eight attention heads. However, the feature embedder agent was allowed to adjust the structural hyperparameters to obtain shallower or deeper MLPs, multiple transformer widths and depths, and multiple two-tower capacity (**Supplementary Data 1-2**). The adjustable hyperparameters included hidden-layer widths for MLPs and two-tower models, transformer dimensionality, number of heads, feed-forward width, and encoder depth for transformer-based models, as well two-tower output dimensionality (**Supplementary Data 1-2**). The candidate model specifications were merged with an internal baseline library and a balanced family-sampling procedure to ensure that no single architecture was the sole candidate during the search. Each candidate model was then trained with the AdamW[47] optimization algorithm under its configurable learning rate (with the default being $10^{-3}$), weight decay, batch size, epoch budget, tokenization and pooling strategies, normalization, symmetry handling, optional contrastive training settings, and regularization settings (**Supplementary Data 1-2**). The final classifier and associated functions were selected after the runs by the fifth and final AI agent, the executor agent (**Figure 1**), based on validation performance under developer-specified metrics, with the Matthews correlation coefficient (MCC[48,49] as the default. Here, the metrics considered by the AI agent included accuracy, F1-score, specificity, and MCC (**Figures 3-5**). In particular, accuracy determines the frequency of correct predictions[49]. F1-score combines the ability of a ML model to identify true positives (recall) and the ability of a ML to make correct positive predictions among all positive predictions (precision) into a single score to determine the ability of a model to identify correct positive labels[49]. Specificity measures the ability of a ML model to correctly identify negative labels[49]. Lastly, MCC represents the correlation between the predicted and actual labels, making it the most comprehensive metrics in evaluating the performance of a ML model[48,49].

$$Accuracy = \frac{TP + TN}{TP + FP + TN + FN} \tag{1}$$

$$Recall = \frac{TP}{TP + FN} \tag{2}$$

$$Precision = \frac{TP}{TP + FP} \tag{3}$$

$$F1 - score = \frac{2 \times Recall \times Precision}{Recall + Precision} \tag{4}$$

$$Specificity = \frac{TN}{TN + FP} \tag{5}$$

$$MCC = \frac{TP \times TN - FP \times FN}{\sqrt{(TP + FP) \times (TP + FN) \times (TN + FP) \times (TN + FN)}} \tag{6}$$

The embedded features of amino acid sequences and candidate ML specifications were all provided to the final AI agent in the platform, the executor, to carry out protein-disjoint training and SHAP[21]-interpretation of the ML predictions (**Figure 1**). Regarding the datasets, the executor agent employed protein-disjoint three-way splitting, as described above, followed by fixed-test cross-fold execution (**Figures 4-5**) by fixing one test set and repeatedly varying the validation set across folds while rebuilding the training dataset each time to eliminate data leakages and improve the generalizability of the ML models on unseen pairs. Furthermore, several other safeguards were included in the training and evaluation logic used by the executor agent to ensure thorough and stringent evaluations of the candidate ML specifications, including class weighting, optional hard-negative mining, optional positive-unlabeled (PU)-aware downweighting, early stopping on validation performance, threshold tuning on validation predictions, optional temperature scaling, bagging, and ensemble aggregation. In particular, class weighting adjusts the contribution of positive and negative data in the loss functions so that any imbalance or uncertain negatives do not dominate optimization[50,51]. Meanwhile, optional hard-negative mining increases attention to negatives that are predicted to be positives to focus the ML on the most confusing non-interacting pairs[52]. On the other hand, optional PU-aware downweighting reduces the penalty assigned to negatives that strongly resemble positives as such pairs may be unrealized positives due to data limitation in PPI[53-56]. By imposing threshold tuning on validation predictions, the executor agent ensured that the final decision cutoff was chosen on held-out validation scores rather than fixed at 0.5, which was necessary to safeguard against class imbalances, shifted score distributions, or the most important metric being MCC rather than accuracy[57]. Optional temperature scaling refers to a post hoc calibration fit on validation logics to fix confidence miscalibration without retraining the entire ML model[58]. Lastly, bagging refers to the practice of training multiple seed-level replicas and averaging their predictions to reduce variance[59,60], while ensemble aggregation is defined as combining predictions across folds and across several top-ranked specifications to improve robustness beyond any single model[61,62].

At the end of the training, the executor agent performed SHAP calculations to identify the important features and feature groups used by the predictive ML models in their predictions (**Figure 1**, **Table 1**, and **Supplementary Data 1-2**). In particular, SHAP is a game-theoretic approach for quantifying the contribution of each feature to the output of ML models[21]. Feedback regarding the testing and validation metrics along with features used by ML models for predictions were provided back to the feature embedder and model designer agents for further improvements (**Figure 1**).

### *Agentic AI platform for rule induction of human-human and virus-human PPI*

For cross-checking and ML interpretability, we modified the agentic AI platform for autonomous training of predictive ML models for PPI (**Figure 1**) to build an agentic AI platform for rule induction of PPI (**Figure 2**). In particular, the agentic AI platform for rule induction of PPI was made of four AI agents with specified roles, with three taken directly from the previous agentic AI platform, including the data collector, data verifier, and feature embedder agents (**Figures 1-2**). However, the model designer and executor agents in the autonomous platform (**Figure 1**) was replaced with a single agent, namely the rule induction agent, in the agentic AI platform for rule induction of PPI (**Figure 2**).

Instead of relying on gradient-based ML models, the rule induction agent rather operated on stable scalar signals derived from pre-trained embeddings, sequence autocovariance descriptors, compartment compatibility, pathway or domain overlaps, domain-domain interaction priors, and graph-neighborhood contexts. It should be noted here that the rule induction agent did not directly consume the entire high-dimensionality vector representations from the feature embedder agent but rather compressed the tensor into scalar summaries, including cosine similarity, dot product, Euclidean distance, mean absolute difference, (*proteinA* and *proteinB*) branch norms, and component-specific similarities for the embedding and descriptor blocks. In fact, the features considered by the rule induction agents could be grouped into four main quantities, including the localization quantities, annotation-overlap quantities, sequence-length or composition-derived quantities, and graph-neighborhood quantities (**Table 2** and **Supplementary Data 3-4**). In particular, localization quantities referred to coarse compartment features derived from protein annotations, including the localization of the two proteins (nucleus, cytoplasm, mitochondrion, endoplasmic reticulum, endosome, extracellular space, membrane, and other related compartments) together with pairwise indicators such as compartment overlap or disjoints. On the other hand, annotation-overlap quantities could be defined as overlap statistics drawn from curated biological knowledge bases, including Pfam[63,64] domain families, Gene Ontology (GO)[65] biological process, molecular function, and cellular component terms, as well as Reactome[66] and domain-domain interaction priors, represented by Jaccard overlap measures[67]. Meanwhile, sequence-length or composition-derived quantities included features such as per-protein length, length differences and ratios, as well as scalar summaries inherited from physicochemical descriptors. Lastly, graph-neighborhood quantities were network-derived properties, including node degrees, common neighbors, neighborhood Jaccard, and preferential attachments. Notably, graph-neighborhood features were built from training positives only to limit data leakage.

The goal of the rule induction agent was to come up with an explicit ruleset in the form of:

$$logit(pair) = bias + \sum weight_i \times f_i(pair\ signals) \tag{7}$$

where $f_i$ was an interpretable indicator, hinge, or linear term. Two complementary search strategies, including the greedy[68] and sparse logistic rule induction[69], were considered by the AI agent in building the rule set. In particular, upon relying on the greedy search strategy, the AI agent performed deterministic forward selection over candidate threshold rules generated from the training distribution, adding a single rule at each step that improved the validation MCC the most and stopping when no further addition resulted in improvements[68]. The advantage of the greedy rule induction approach was to produce compact and transparent rules[68] (**Table 2**). However, the rule induction agent might miss rules that were only useful when combined[68]. By contrast, the sparse logistic rule induction initially constructed a much larger candidate matrix that were capable of inducing indicator rules, hinge rules, direct linear terms, and conjunctions[69]. Afterwards, the sparse logistic rule induction fit an L1-penalized logistic model so that only a sparse subset of candidate rules retained non-zero weights[69]. The advantage of the sparse logistic rule induction was the ability to recover multicomponent rules, but the produced ruleset was far from compact (**Table 2**).

Furthermore, the rule induction supported task-aware exclusions as one harmless feature in one biological regime could become a shortcut in another[69], and missingness-proxy features could be suppressed, union-overlap rules could be capped if they became highly broad, and threshold selection was tuned on validation outputs. In particular, missingness-proxy features could be suppressed in case variables such as whether a protein had localization annotation could only reflect annotation completeness or study bias rather than interaction biology[70], especially in the case of virus-human PPI where viral data were much more limited. Furthermore, union-overlap rules were capped to ensure that overly large annotation unions associated with one side of the interaction pairs did not turn into a diffuse, non-discriminative shortcut rather than meaningful biological constraints[71], also in the case of virus-human PPI.

The exported rulesets for PPI by the rule induction agent could be cross-checked against the features found by SHAP[21] to be important for PPI. Agreement between the rulesets and SHAP[21]-identified features would increase the confidence that the platforms were learning true biological laws rather than exploiting hidden artifacts.

## Results

We instructed a coding agent, based on ChatGPT 5.2 and 5.4 Thinking, to construct two agentic AI platforms for autonomous training of predictive ML models and for rule induction of human-human and virus-human PPI, to examine the abilities of agentic AI in training ML models and in understanding biological sciences. The logic behind the construction of the agentic AI platforms was based on the processes of human developers training ML models and understanding rules governing biological processes. As such, the agentic AI platform for autonomous training of predictive ML models for PPI comprised five AI agents with distinct roles: data collector, data verifier, feature embedder, model designer, and executor (**Figure 1**). Meanwhile, the agentic AI platform for PPI rule induction included four AI agents, retaining the first three agents from the agentic AI platform for autonomous training and replacing the model designer and executor with the rule induction agent (**Figure 2**). All AI agents were based on OpenAI API gpt-5.4 models. At the end of the training, SHAP[21]-identified features (from the agentic AI platform for autonomous training) (**Table 1**) were compared with the explicit rulesets (from the agentic AI platform for rule induction) (**Table 2**) to determine whether the ML models truly learned the biological rules governing PPI or were relying on hidden artifacts for making the predictions.

### *Predictive ML models for human-human and virus-human PPI trained by agentic AI*

The data collector agent curated a total of 96,637 positive and 96,623 negative human-human interactome data, sampled from 42,547 reviewed human proteins (those that were manually annotated and reviewed by UniProt curators), primarily from the IntAct[22] database, Richoux et al.[25,26], Negatome[27,28], and compartment-based plus CORUM[29]-informed co-complex filtering (**Supplementary Data 1** and **3**). We intentionally forced the data collection agent to use a 1:1 ratio for positive and negative data to avoid inflation of metrics. The feature embedder employed the ESM-2 PLM[38], particularly the *facebook/esm2_t33_650M_UR50D*, to serve as the encoder for protein amino acid sequences (**Supplementary Data 1**). The model designer and executor agents determined that the two-tower was the best ML model for predictions of human-human PPI, with all related ML specifications included in **Supplementary Figure 1** and **Supplementary Data 1**. The overall metrics from the training, testing, and validation sets for the predictive ML model for human-human PPI were included in **Figure 3A** and **Supplementary Data 1**, while the metrics from training, testing, and validation sets during the three-way protein-disjoint cross-fold execution for the predictive ML model for human-human PPI were shown in **Figure 4** and **Supplementary Data 1**. Under the protein-disjoint cross-fold execution, the final ensemble achieved a testing accuracy of 87.3%, a testing F1-score of 87.1%, a testing specificity of 89.1%, and a testing MCC of 74.6% (**Figure 4E** and **Supplementary Data 1**), demonstrating that the trained ML model was generalizable to unseen proteins. Here, the good performance of the predictive ML model for human-human PPI could also be attributed to the relative abundance of curated positive and negative data for human-human PPI, allowing for the use of reviewed-only proteins (those that were manually annotated and reviewed by UniProt curators), IntAct[22] confidence screening, Negatome[27,28]-assisted, compartment-disjoint, and CORUM[29]-based co-complex negative cleaning.

On the other hand, the data collector agent assembled 47,652 positive and 47,652 negative virus-human interactome data, sampled from 2,110 viral proteins and 105,657 human proteins, primarily from HVIDB[30] and VirusMentha[31] for the positive data (**Supplementary Data 2** and **4**). The negative data for virus-human PPI were mostly synthetic, due to the scarcity of explicit negative evidence in pathogen-host interaction data, resulting in noisier, lower-confidence negative datasets. Here, a 1:1 ratio was also enforced on positive and negative data to avoid metric inflation if the ML models learned nothing and predicted everything as negative. The feature embedder used the same ESM-2 PLM[38] (*facebook/esm2_t33_650M_UR50D*) as the human-human PPI to encode the virus-human PPI (**Supplementary Data 2**). Here, the two-tower ML model was again selected by the model designer and executor agents to be the best ML architecture for predicting virus-human PPI (**Supplementary Figure 1**), with all associated ML specifications included in **Supplementary Data 2**. The overall metrics from the training, testing, and validation sets for the predictive ML model for virus-human PPI were included in **Figure 3B** and **Supplementary Data 2**, while the metrics from training, testing, and validation sets during the three-way protein-disjoint cross-fold execution for the predictive ML model for virus-human PPI were shown in **Figure 5** and **Supplementary Data 2**. The final ensemble under the protein-disjoint cross-fold execution achieved a testing accuracy of 86.5%, a F1 score of 87.0%, a specificity of 85.6%, and an MCC of 73.0% (**Figure 5E** and **Supplementary Data 2**). Despite limited data availability for virus-human PPI, the predictive ML model trained by agentic AI showed robust performance on unseen proteins with optional temperature scaling enabled (**Supplementary Data 2**).

SHAP[21] analyses of the predictions made by the predictive ML models for human-human and virus-human PPIs trained by agentic AI showed that the ML models relied heavily on PLM-derived embeddings for classification (**Table 1** and **Supplementary Data 1-2**). In particular, the SHAP-determined group importances were found in the tokens and embeddings of both human proteins, with much smaller contributions from the Z-ACC[40-42] and E-ACC[39,41,42], in the human-human PPI (**Table 1** and **Supplementary Data 1**). In the virus-human PPI, the SHAP[21]-determined group importances shifted strongly towards the tokens and embeddings of human proteins, with smaller contributions from the tokens and embeddings of viral proteins and marginal contributions from the Z-ACC[40-42] and E-ACC[39,41,42] (**Table 1** and **Supplementary Data 2**). This was consistent with the fact that the human protein data were more abundant in the virus-human PPI, with only 2,110 viral proteins sampled compared to 105,667 human proteins (**Supplementary Data 2** and **4**). Here, it should be noted that no contribution was recorded from the contrast and concordance channel in the predictive ML models for human-human and virus-human PPI, meaning the features from the two proteins were considered separately for classification, likely due to the use of two-tower ML architecture in training the ML models (**Table 1, Supplementary Figure 1,** and **Supplementary Data 1-2**).

***Rule induction of human-human and virus-human PPI by agentic AI***

The agentic AI for rule induction used identical datasets curated for human-human and virus-human PPI as the agentic AI for autonomous training of predictive ML models (**Supplementary Data 3** and **4**). Here, to ensure that the biological features induced by agentic AI for PPI were actually important, we forced the feature embedder agent to use slightly different PLM encoders, with the *facebook/esm2_t12_35M_UR50D*[38] used as the encoder for human-human protein amino acid sequences (**Supplementary Data 3**) and the *facebook/esm2_t30_150M_UR50D*[38] used as the encoder for virus-human protein amino acid sequences (**Supplementary Data 4**).

The rule induction agent used a hybrid search strategy for rule induction of human-human PPI, allowing it to compare greedy[68] forward selection with sparse logistic[69] rule induction and retaining the ruleset with the better validation MCC (**Supplementary Data 3**). On the three-way protein-disjoint split for human-human PPI, the induced ruleset by the AI agent achieved a training MCC of 81.6%, validation MCC of 81.2%, and testing MCC of 81.0%, with the testing accuracy of 89.6%, F1-score of 88.5%, and specificity of 99.9% (**Figure 3C** and **Supplementary Data 3**). The high specificity indicated a conservative and compact ruleset (**Figure 3C** and **Supplementary Data 3**). In fact, the final ruleset for human-human PPI contained only two active rules, with a tuned decision threshold of 0.51, starting from 72 candidate signal dimensions (**Table 2** and **Supplementary Data 3**). The two active rules for human-human PPI induced by the AI agent included *comp_disjoint_known < 1*, which rewarded interacting pairs that were not known to occupy disjoint subcellular compartments, and *pfam_jaccard > 0*, which favored shared domains and families for interacting pairs (**Table 2** and **Supplementary Data 3**). Taken together, the rules induced by AI agents suggested that human proteins were more likely to interact when they were not spatially distant and when they shared domain or family organizations[72,73]. Notably, the fact that a two-rule system was generalizable on unseen proteins in the three-way protein-disjoint split for human-human PPI suggested that a large number of human-human PPI were governed by domain compatibility and spatial co-accessibility.

On the other hand, the rule induction agent employed the sparse logistic[69] rule induction for the virus-human PPI to search for a larger candidate space, with rule caps up to 60 (**Supplementary Data 4**). On the three-way protein-disjoint split for virus-human PPI, the induced ruleset by the AI agent achieved a training MCC of 74.8%, validation MCC of 72.3%, and testing MCC of 75.4%, with a testing accuracy of 87.7%, F1-score of 87.5%, and specificity of 90.1% (**Figure 3D** and **Supplementary Data 4**). The final rules for virus-human PPI appeared to generalize reasonably well on unseen proteins, with 60 weighted rules retained using a tuned decision threshold of 0.4 (**Supplementary Data 4**). Inspection of the final rules induced by AI agent illustrated three dominant themes to explain virus-human PPI (**Table 2** and **Supplementary Data 4**). First, compartment and localization features of human proteins dominate the rules, with positive rules rewarding human proteins in the mitochondrion, endoplasmic reticulum, endosome, extracellular space, and selected nuclear compartments (**Table 2** and **Supplementary Data 4**). Second, signals derived from the embedding of human proteins were important, with interacting pairs favoring low mean absolute pairwise differences and penalties for large embedding norms (**Table 2** and **Supplementary Data 4**). Third, interacting pairs exhibit a smaller average feature-space separation between viral and human proteins (**Table 2** and **Supplementary Data 4**). Taken together, the final rules induced by agentic AI described the virus-human PPI as a dominant human host-protein problem in which where the human proteins resided and how their representations were positioned in the feature space, relative to the viral proteins, mattered substantially[16,74-76].

***Cross-checking of SHAP-identified features and biological rules induced for human-human and virus-human PPI by agentic AI***

Overall, the SHAP[21]-identified features and induced rules for human-human and virus-human PPI appeared to converge biologically, even with slightly different ESM-2 PLM encoders[38] (**Supplementary Data 1-4**). In particular, for the human-human PPI, the SHAP[21]-identified features mostly included the tokens and embeddings of the human proteins (**Table 1** and **Supplementary Data 1**), while the ruleset induced by agentic AI focused on co-accessibility and domain or family overlap (**Table 2** and **Supplementary Data 3**). Therefore, the SHAP[21]-identified features and rules induced by the AI agent were consistent, as they both captured the separate features of both human proteins to be crucial for human-human PPI.

Similarly, the same could be said for the virus-human PPI. In particular, the SHAP[21] analysis revealed the dominance of tokens and embeddings from human proteins in predicting virus-human

PPI (**Table 2** and **Supplementary Data 2**), whereas the ruleset induced by agentic AI highlighted the importance of human protein localization and feature constraints (**Table 2** and **Supplementary Data 4**). However, the ruleset induced by agentic AI for virus-human PPI contained a larger proportion of compartment and localization terms (**Table 2** and **Supplementary Data 4**). Even so, both SHAP[21] and induced rules supported the interpretation that human protein features were the most crucial factor in determining virus-human PPI.

Taken together, the rule-induction results could serve as cross-checks for the predictive ML models. However, direct comparisons should be interpreted cautiously, as the inference layers differed between the predictive ML models and the rule induction agents. Furthermore, the predictive ML models were trained and inferred in an ensemble across multiple folds (varying training, testing, and validation sets to assess generalization) and ML specifications. Nevertheless, the cross-checks between results from predictive ML models and rules induced by AI agents confirmed that the agentic AI workflows could identify intelligible biological regularities in PPI rather than opaque ML artifacts.

## Discussion

In this work, we have evaluated the ability of AI agents to follow instructions and construct agentic AI workflows that could autonomously train predictive ML models and induce rulesets governing human-human and virus-human PPI (**Figures 1-2**). Here, a coding agent received high-level instructions from human developers and coordinated data collection, identifier harmonization, negative-dataset design, sequence embedding, model design, calibration, and execution or rule induction under three-way protein-disjoint evaluations (**Figures 1-2**). Our work demonstrated that agentic AI could be used to perform the entire typical protocol of training, validation, and explanation in ML reproducibly.

For the human-human PPI datasets, we demonstrated the ability of AI agents to achieve strong dataset hygiene by imposing filtering of unreviewed proteins (those that have not been manually annotated and/or reviewed by UniProt curators), Negatome[27,28]-aware negatives, and co-complex and compartment-aware exclusions for synthetic negative data. However, the AI agents appeared to struggle where human developers struggled with the datasets for virus-human PPI as data related to proteins are more limited and heterogeneous compared to human proteins. However, the predictive ML models and ruleset induced by agentic AI appeared robust as they showed reasonable performances (**Figures 3-5**) on three-way protein-disjoint executions.

For the predictive ML models, the agentic AI selected the two-tower as the primary ML architectures for classifying human-human and virus-human PPI (**Figure 1, Supplementary Figure 1,** and **Supplementary Data 1-2**). The deployed two-tower scorers consumed the feature blocks of the two proteins directly, while keeping the representations of the protein separate and inference scalable. Furthermore, the agentic AI decided to incorporate several additional safeguards, including class weighting, temperature scaling, optional hard-negative mining, and three-way protein-disjoint cross-fold validation, to ensure robust training and generalization to unseen proteins.

The agentic AI for rule induction was built to determine whether the features employed by the predictive ML for PPI could be converted into interpretable logics and whether AI agents could understand the biological rules governing PPI (**Figure 2** and **Supplementary Data 3-4**). Overall, the SHAP[21]-identified features from the predictive ML models and rulesets induced by agentic AI appeared to converge on biological context (**Tables 1-2** and **Supplementary Data 1-4**). For human-human PPI, both SHAP[21] and the ruleset identified compartment and domain-family compatibility as governing a substantial fraction of human-human interactomes[72,73](**Tables 1-2** and **Supplementary Data 1, 3**). Meanwhile, the virus-human PPI was governed by a broader ruleset, drawing on human protein localization, human protein annotation context, human protein graph-neighborhood structure, and sequence-derived scalar contrasts[16,74-76] (**Tables 1-2** and **Supplementary Data 2, 4**). In fact, the ruleset for virus-human PPI showed that virus-human recognition was regulated through multiple mechanistic routes rather than one universal motif[16,74-76]. Therefore, the rule induction platform served as a substantive cross-check for the predictive ML platform through both their agreements and differences.

While the agentic AI showed good performances in predicting PPI in this work, it should be noted here that the predictions were made entirely on curated published data rather than on unpublished datasets. Therefore, it would be necessary to validate the trained predictive ML models and rulesets induced by AI agents on external unpublished data to ensure their rigorous performance. Furthermore, although the feature embedder was designed to extract and embed structural and functional features, this first study of agentic AI for PPI relied entirely on amino acid sequences to predict PPIs. For future studies, we will collect and build structural models[77-79] to further improve the performance of the agentic AI workflows.

In conclusion, we have demonstrated the full agentic orchestration from data planning to execution and rule induction to explanation for human-human and virus-human PPI. A coding agent was used to build two interpretable agentic AI platforms for PPI. The first agentic AI autonomously assembled data, trained and evaluated on three-way protein-disjoint cross-fold executions, as well as explained two-tower predictive ML models for PPI. On the other hand, the second agentic AI induced explicit biological rules that governed PPI. Taken together, the two agentic AI platforms demonstrated that agentic AI could support both accurate PPI prediction and mechanistically meaningful rule discovery, opening the door to future applications across various areas.

**Data Availability Statement**

Data supporting the findings of this study are included in the article and its Supplementary Data files, which can be found on Zenodo (DOI: 10.5281/zenodo.19701990).

**Code Availability Statement**

The study extensively used ChatGPT 5.2 and 5.4 Thinking plus OpenAI API agents based on gpt-5.4 model in generating codes and analyzing data. The generated scripts will be made available upon approval from funding sponsor.

**Author Contributions**

H.N.D. designed and performed research, analyzed data, and wrote manuscript. J.Z.K. acquired funding, supervised research, and wrote manuscript. O.A.N. acquired funding, supervised research, and wrote manuscript. S.G. acquired funding, supervised research, analyzed data, and wrote manuscript. All authors contributed to the final version of the manuscript.

**Conflict of Interest Statement**

The authors declare no competing interests.

## Acknowledgements

This work was supported by the Joint Science and Technology Office, Defense Threat Reduction Agency under the Rapid Assessment of Platform Technologies to Expedite Response (RAPTER) program (award no. HDTRA1242031). This study was also supported through funding from the Sandia National Laboratory Directed Research and Development (LDRD) program. Sandia National Laboratories is a multi-mission laboratory managed and operated by National Technology and Engineering Solutions of Sandia, LLC, a wholly owned subsidiary of Honeywell International Inc, for the U.S. Department of Energy's National Nuclear Security Administration under contract DE-NA0003525. The authors would like to thank the computational resources provided by the Los Alamos National Laboratory Institutional Computing and AI portal. The authors thank Dr. Traci Pals and Dr. Bob Webb for their support of this work. The views expressed in this article are those of the authors and do not reflect the official policy or position of the U.S. Government.

**Figures**

**Figure 1. Scheme of the agentic AI platform for autonomous training of predictive ML models for human-human and virus-human protein-protein interactions (PPI).** A coding agent received high-level instructions from the human developer to generate the necessary codes to build the agentic AI platform. The agentic AI platform consisted of five AI agents. The first AI agent, data collector, integrated data from reliable publications, manually curated datasets, and databases to build the positive and negative datasets and split the data into training, testing, and validation sets using the ratio of 70:20:10 by proteins rather than interaction pairs. The datasets were then provided to the second AI agent, data verifier, to interrogate class balance, possible leakage risk, negative inflation, and possible data hallucinations. Feedback was provided to the data collector to force rebuilding of the datasets if any violations were detected. Afterwards, the verified datasets were supplied to the third AI agent, feature embedder, to extract relevant descriptors and create proper embeddings, which were then supplied into the fourth agent, model designer. Given the embeddings, the model designer agent then selected the proper ML architecture along with other ML specifications, including activation and loss functions. Finally, the fifth agent carried out the training and evaluation of the predictive ML model for PPI, including determining which features were used by the ML model to make the predictions using the Shapley additive explanation (SHAP) approach. Feedback was then provided to the third and fourth agents to find strategies to improve the predictive power of the ML model.

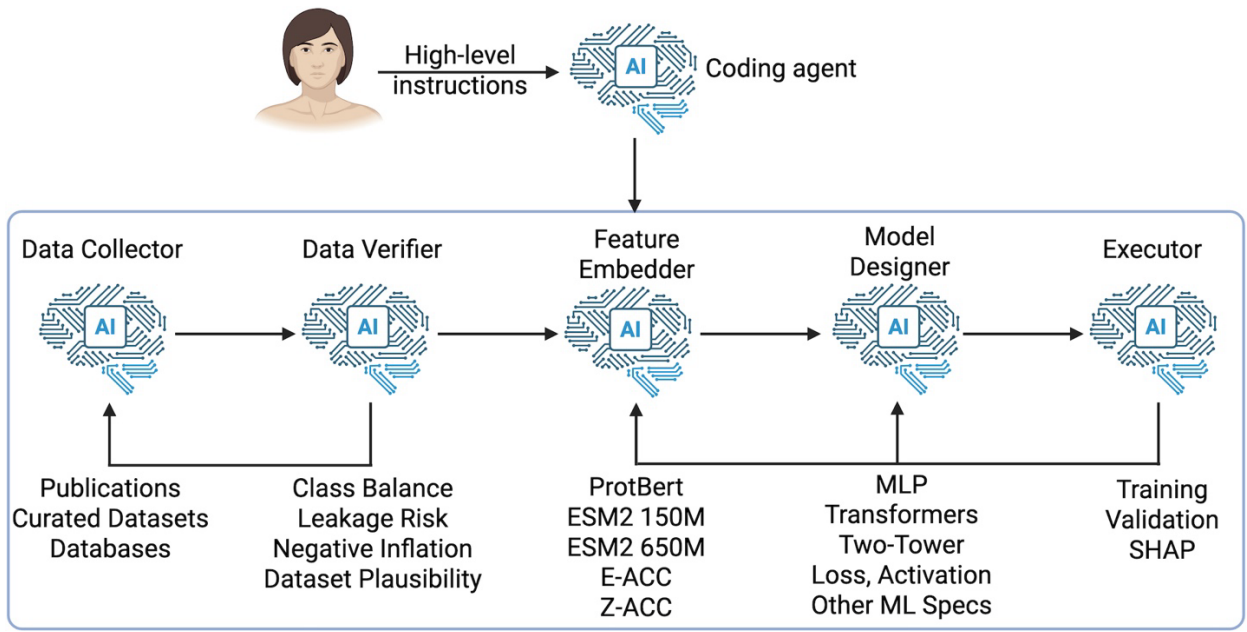

**Figure 2. Scheme of the agentic AI platform for rule induction of human-human and virus-human PPI.** The agentic AI platform for rule induction was developed from the platform for autonomous training of predictive ML model and retained similar AI agents of coding, data collection, data verifier, and feature embedder. However, the model designer and executor agents were replaced by the rule induction agent, which took in the embeddings by the feature embedder to induce an explicit ruleset for PPI.

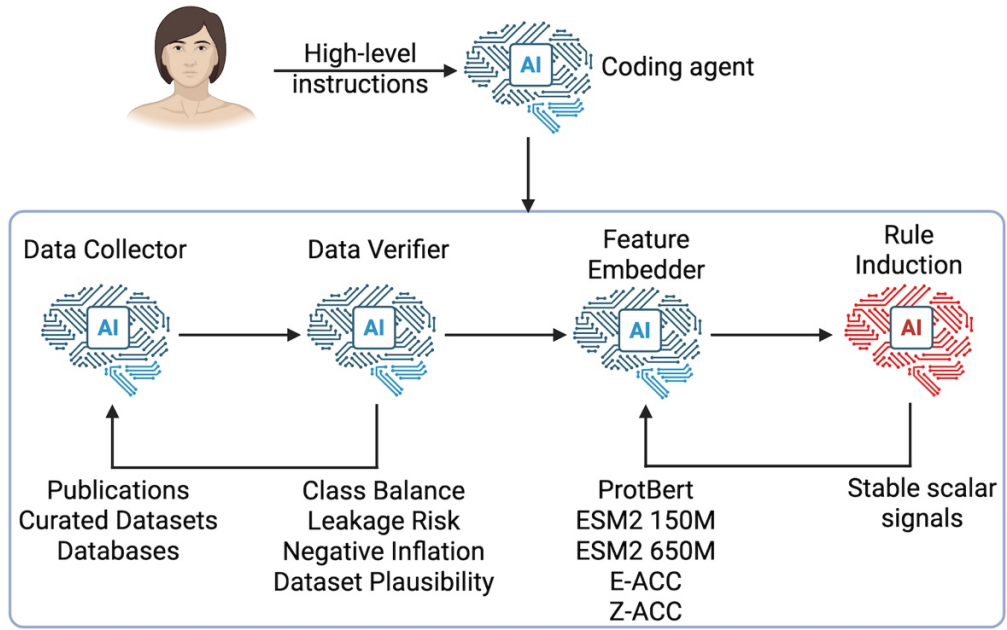

**Figure 3. Performance metrics for predictions made by the agentic AI platforms for autonomous training of predictive ML model and rule induction of human-human and virus-human PPI on the training, testing, and validation datasets. (A-B)** Performance metrics of the predictive ML models trained by the agentic AI for autonomous training on the training, testing, and validation datasets for human-human **(A)** and virus-human PPI **(B)**. **(C-D)** Performance metrics of the agentic AI for rule induction of human-human **(C)** and virus-human PPI **(D)** on the respective training, testing, and validation datasets. Four metrics were considered, including accuracy, F1 score, specificity, and Matthews correlation coefficient. The metrics for training datasets were colored green, for testing were colored blue, and for orange were colored orange.

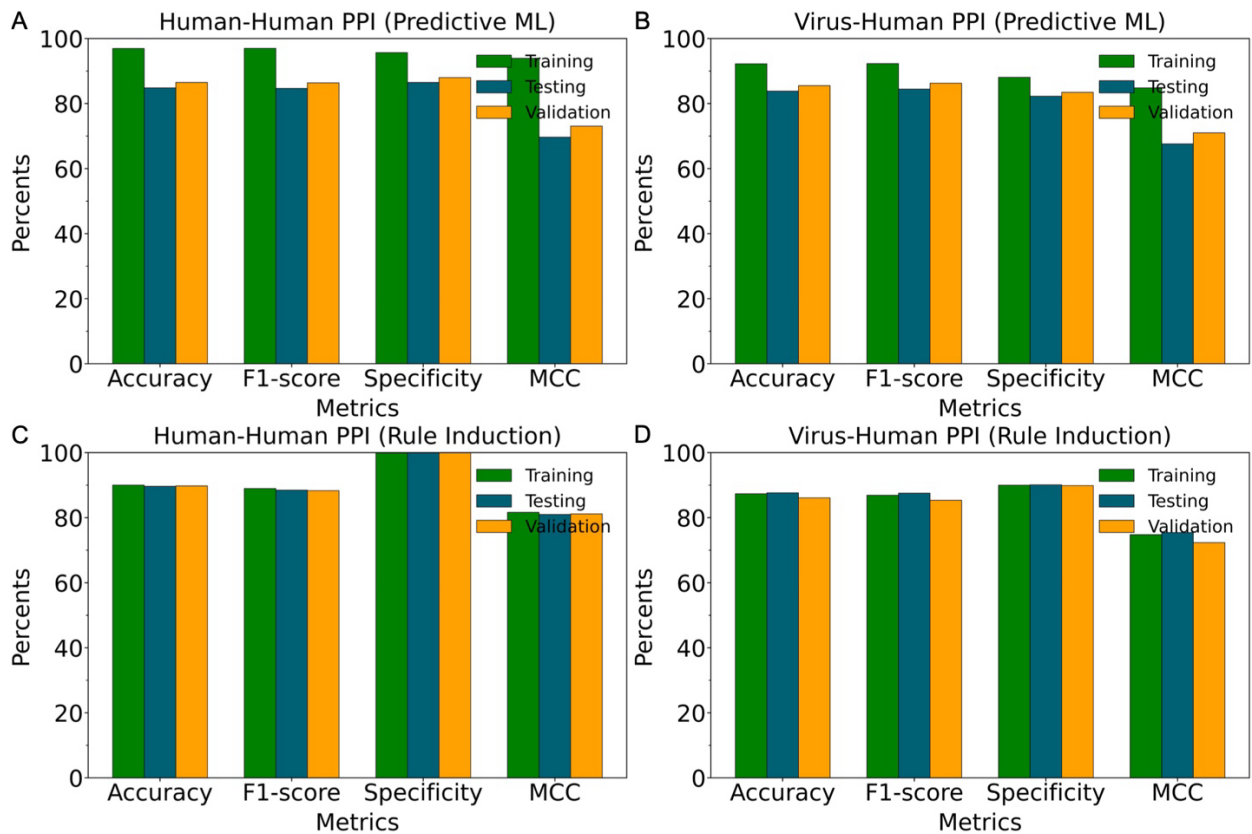

**Figure 4. Performance metrics for predictions made by predictive ML models trained by the agentic AI platform for autonomous training on five different ensembles of training, testing, and validation datasets for human-human PPI.** Four metrics were considered, including accuracy, F1 score, specificity, and Matthews correlation coefficient. The metrics for training datasets were colored green, for testing were colored blue, and for orange were colored orange.

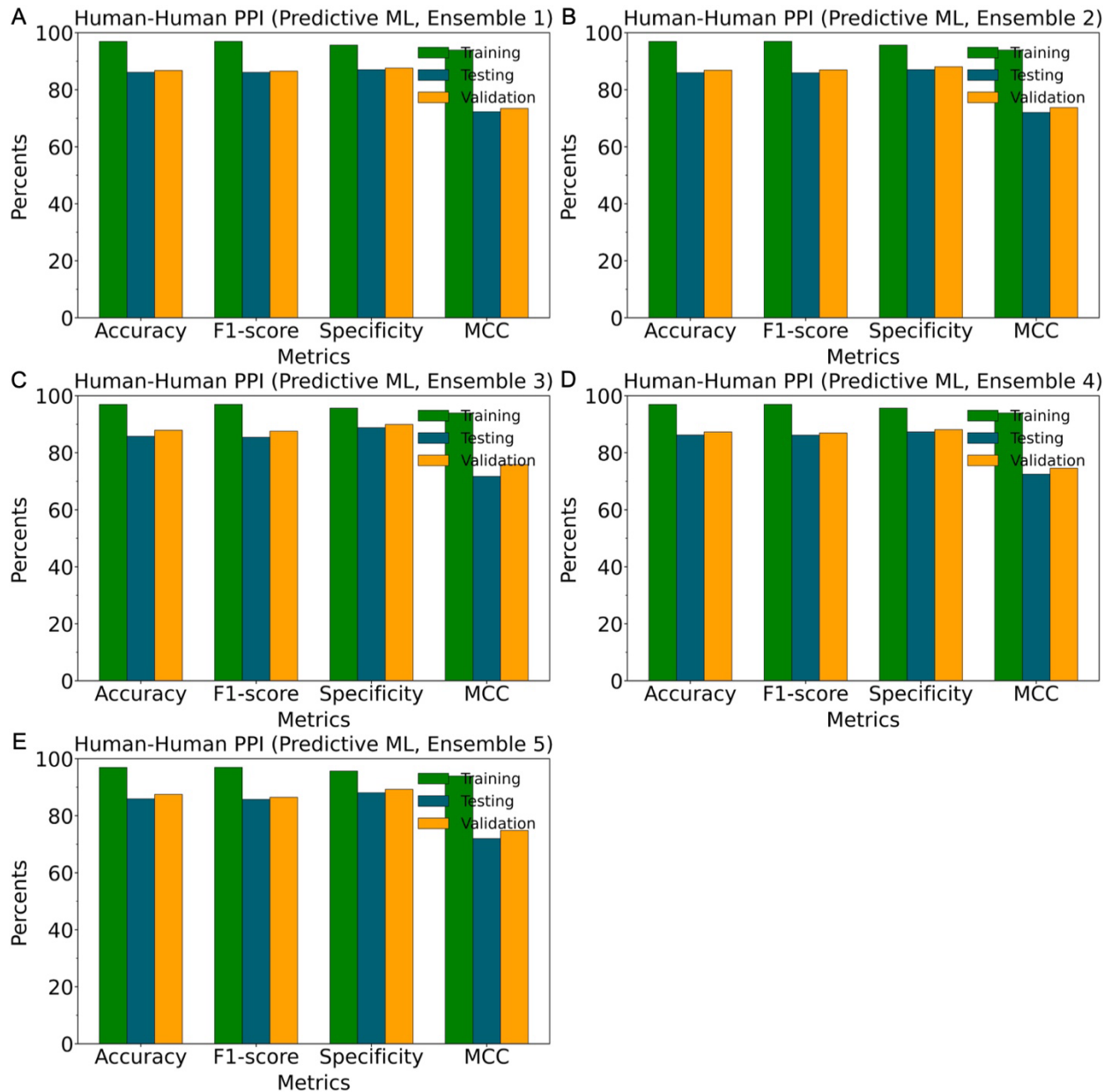

**Figure 5. Performance metrics for predictions made by predictive ML models trained by the agentic AI platform for autonomous training on five different ensembles of training, testing, and validation datasets for virus-human PPI.** Four metrics were considered, including accuracy, F1 score, specificity, and Matthews correlation coefficient. The metrics for training datasets were colored green, for testing were colored blue, and for orange were colored orange.

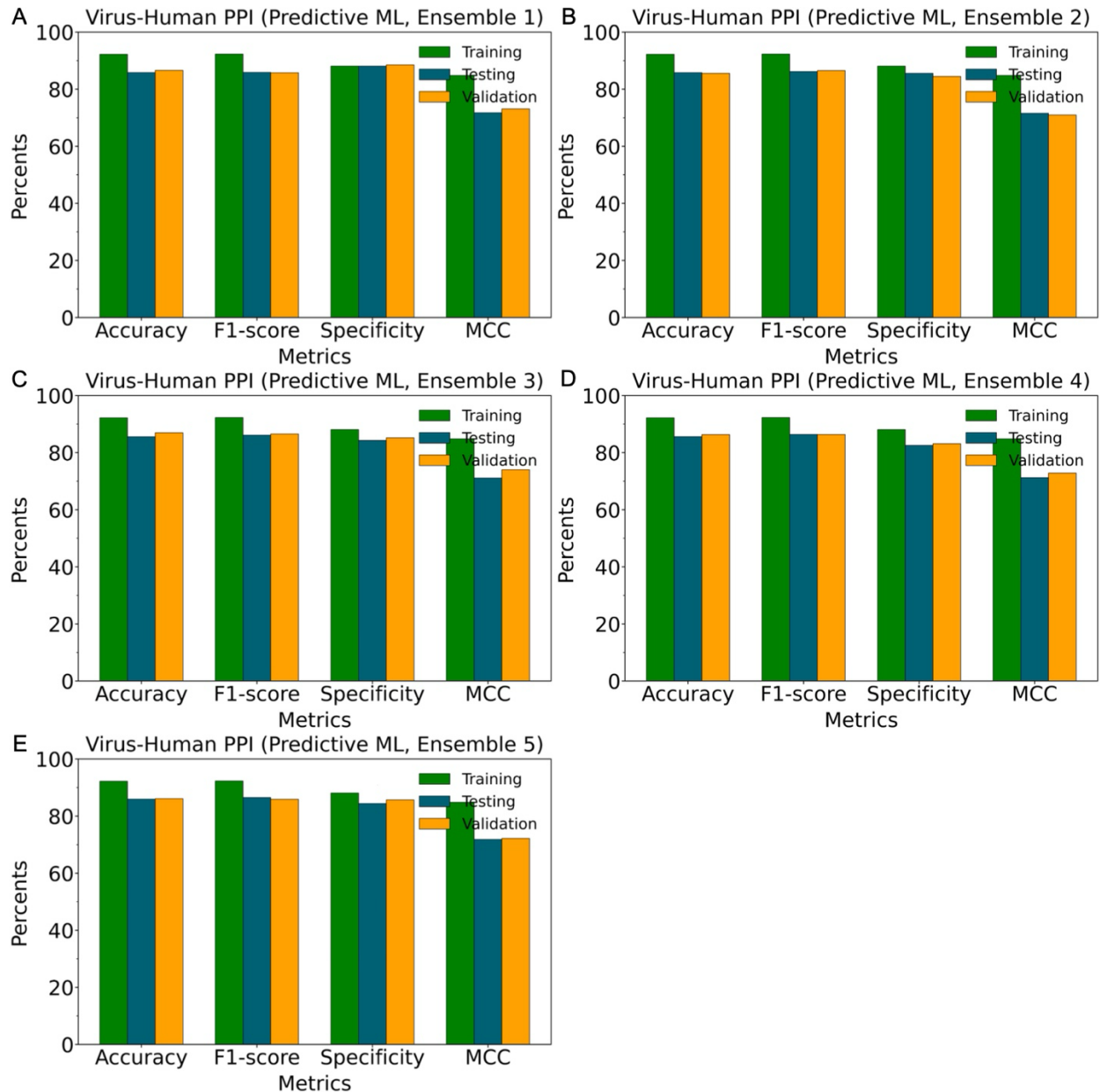

**Table 1. Features determined by the Shapley additive explanation (SHAP) approach to be important for the predictions made by the predictive ML models for human-human and virus-human PPI.** Here, Z-ACC[40-42] refers to the lagged auto-covariance transformation of the five-dimensional Sandberg z-scale representation for amino acids. E-ACC[39,41,42] refers to the lagged auto-covariance transformation of the one-dimensional Eisenberg consensus hydrophobicity scale for amino acids.

| Task | Dominant Groups | Interpretations |
|---|---|---|
| Human-human PPI | - Embeddings and tokens from both human proteins dominated.<br>- Z-ACC and E-ACC made minor contributions. | The ML model mostly used sequence-embedding similarity across both human proteins for predictions of human-human PPI. |
| Virus-human PPI | - Embeddings and tokens from human proteins dominated over those from viral proteins.<br>- Z-ACC and E-ACC made minor contributions. | In the ML model for predictions of virus-human PPI, human protein representations contributed more deployable signals than the viral protein representations. |

**Table 2. Representative rules induced by the agentic AI platform for human-human and virus-human PPI.**

| Task | Rule | Weight | Interpretation |
|------|------|--------|----------------|
| **Human-human PPI** | comp_disjoint_known < 1 | 1.0 | Pairs were favored to interact when both human proteins were not known to occupy disjoint coarse cellular compartments. |
| | pfam_jaccard > 0 | 1.0 | Pairs were favored to interact when both human proteins shared non-zero Pfam family overlap, consistent with domain-level compatibility. |
| **Virus-human PPI** | absdiff_mean < 0.13 | 3.1 | Pairs were favored to interact when the viral protein and human protein showed small average feature-space separation. |
| | tar_in_mitochondrion > 0 | 1.2 | Human proteins annotated to the mitochondrion were more favored to interactions. |
| | tar_in_er > 0 | 0.9 | Human proteins localized in the endoplasmic reticulum were more favored to virus-human PPI. |
| | tar_in_endosome > 0 or tar_in_extracellular > 0 | 0.55 0.54 | Endosomal and extracellular proteins are more likely to interact with viral proteins. |
| | tar_in_nucleus < 1 and tar_in_cytoplasm < 1 | -1.1 | Human proteins lacking both nucleus and cytoplasm annotations were penalized against virus-human PPI. |
| | human_norm_emb > 7.7 | -0.9 | Extremely large embedding norm of human protein was penalized against virus-human PPI. |

**Supplementary Figure 1. ML specifications employed by the agentic AI for autonomous training of the ML models for predictions of human-human and virus-human PPI.** Image was generated by ChatGPT Image 2.0 based on the outputs generated by the agentic AI.

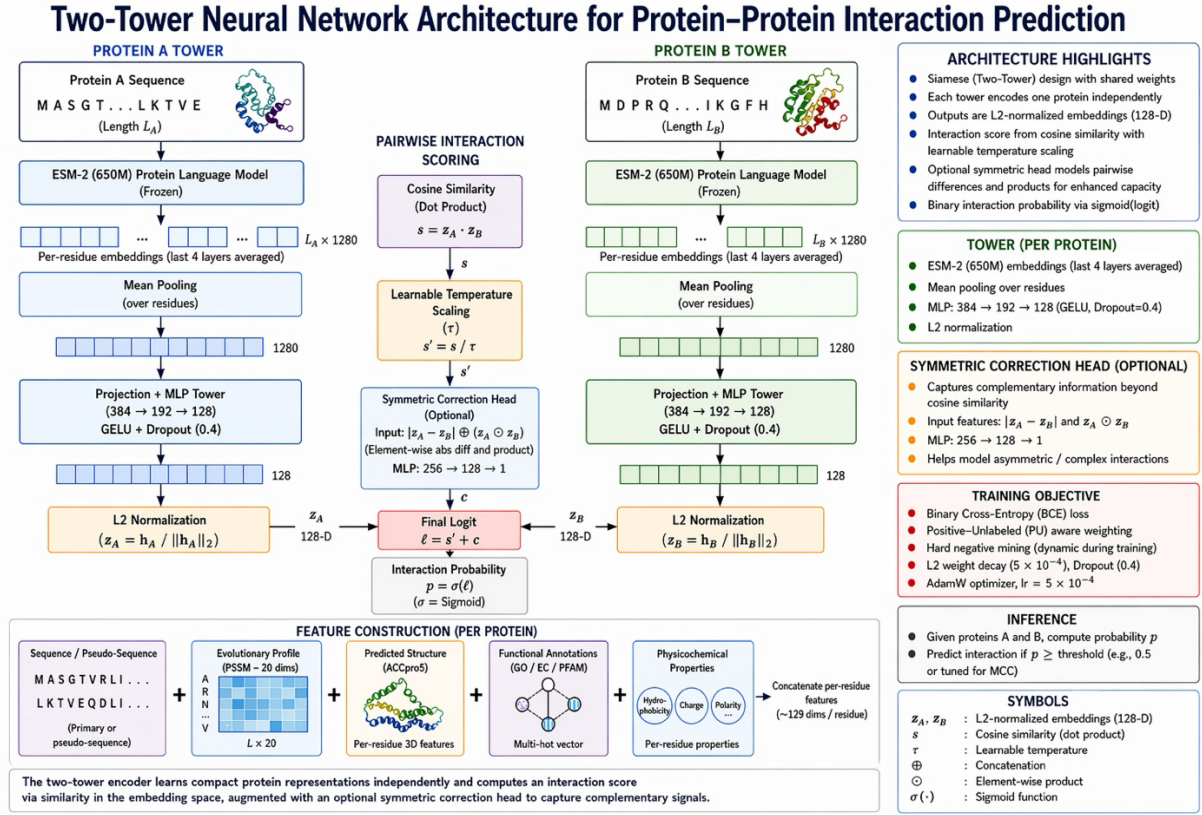